\documentclass[runningheads]{llncs}
\usepackage{graphicx}
\usepackage{todonotes}
\begin{document}
\title{The Danger of Reverse-Engineering of Automated Judicial Decision-Making Systems}

%
%
\author{Masha Medvedeva\inst{1,2}\orcidID{0000-0002-2972-8447} \and
Martijn Wieling\inst{1}\orcidID{0000-0003-0434-1526} \and
Michel Vols\inst{2}\orcidID{0000-0002-5762-8697}}
\authorrunning{M. Medvedeva et al.}
%
\institute{Center for Language and Cognition Groningen, University of Groningen \and
Department of Legal Methods, University of Groningen\\
\email{\{m.medvedeva,m.b.wieling,m.vols\}@rug.nl}}
\maketitle              
\begin{abstract}
In this paper we discuss the implications of using machine learning for judicial decision-making in situations where human rights may be infringed. We argue that the use of such tools in these situations should be limited due to inherent status quo bias and dangers of reverse-engineering. We discuss that these issues already exist in the judicial systems without using machine learning tools, but how introducing them might exacerbate them. 
\end{abstract}
\section{Introduction}

Legal technology is rapidly gaining interest. Many law firms and government organizations use tools to automate, speed up and simplify their work. 
For example, Estonia introduced a robo-judge to resolve small claims in courts \cite{niiler2019can}, and the courts in the USA are using risk-assessment tools to determine the amount of bail \cite{stevenson2018assessing}.

Most of the technology and algorithms used in legal tech were developed in other domains and then applied to legal data. However, in the legal domain, similar to, for instance, the medical domain, the consequences of using these algorithms may be much greater than for the domains for which the algorithms originally were created. Often those consequences are not considered when the algorithms are applied to a new domain.
 
One of the subjects that the field of AI and Law is fascinated with is making a fair and unbiased decision-making system that would be able to assist judges in their decision-making, or even replace them in courts \cite{bex2020juridische}. The idea of making judges' work easier, and the courts faster and more transparent is appealing. 
We agree that automatic systems may be suitable to be used for some levels of judicial decision-making (see \cite{stranieri2006knowledge} and \cite{berk2019machine} for a variety of methods), particularly in low-stakes environments dealing with minor claims. However, we will argue that in situations where human rights may be infringed, 
this may not be a good idea.

Consider the following hypothetical situation in which there is a criminal court that uses a fully automated system to make decisions based on the facts of the case provided by the parties. The model weighs the facts and comes up with the decision based on the precedent. The system is completely transparent and explainable and makes its decisions with very high accuracy. Now consider the perspective from the prosecutor's office that deals with many cases appearing before the court using this system. Having dealt with a very large number of cases, they are able to understand (almost) exactly how the automatic system works and reverse-engineer (i.e.~determine the rules which determine the judgements) it in order to evaluate the outcome when presenting the case to the court in a certain way. Let's assume they are able to reverse-engineer the system in such a way, that their system overlaps with the court's system for 99\%. 
They play out every scenario before the trial, and make sure they use the strategy that (almost) guarantees their win. If they are not able to identify a winning strategy, they decide not to prosecute, thereby providing the office with a 99\% winning rate. Would such a scenario be desirable? What if instead of the prosecutor's office, the system is available to a very large law firm that defends the most violent criminals or corrupt politicians and is able to present the facts in such a way that the system would be favourable to their clients?

We are, of course, quite far from building such a system just yet. Predictive models that are developed today are nowhere near having a high enough accuracy to be used for decision-making in actual courts.\footnote{For example, Chalkidis et al.~\cite{chalkidis-etal-2019-neural} report an F1-score of 80.5 for predicting whether there is going to be any violation in future decisions of the European Court of Human Rights, and Medvedeva et al.~\cite{medvedeva2020using}, analyzing a different subset of the data, report an average F1-score of 0.66  for predicting a violation in the three most-frequent articles of the European Court of Human Rights.} 
Unfortunately that does not stop the courts from using such models. Specifically, decision-predicting systems are now used for decision-making, for example by the US courts. 
The most common system is a risk assessment tool called COMPAS\footnote{http://www.northpointeinc.com/files/downloads/FAQ\_Document.pdf}, which is used to decide on bail and probation, and sometimes is even used for sentencing \cite{angwin2016machine} even though it may show racial bias. Unfortunately, the system predicts recidivism with the same accuracy as a random person with little or no criminal expertise \cite{dressel2018accuracy}.

While the performance of the decision-predicting models is not very high, the interest in using such systems is still growing. Aletras et al.~\cite{aletras2016predicting} suggested that using predictive models could be useful to prioritise certain cases over others in the European Court of Human Rights. Chen et al.~\cite{chen-etal-2019-charge} suggested using their prison-term-prediction system for the Chinese Supreme Court as an anonymous check of the judge's decision, while their system predicts the exact same term as the judge in only about 9\% of the cases. Zhong et al.~\cite{zhong2018legal} suggested that legal judgement predictions can be used to assist lawyers and judges. Their model TopJudge is designed to predict prison terms, but occasionally suggests a death penalty whereas the judge decided on a prison term of less than a year.

In this paper, we argue that in high stake court cases (e.g. cases in which people's lives or fundamental rights are at stake or that may have a strong influence on public policy), the use of such systems may have implications that may be detrimental to peoples' lives and accessibility to justice, and should be prevented.

While previous research has considered several ethical and technical considerations regarding such systems, we specifically focus on the dangers of reverse-engineering. In addition, we discuss several additional reasons why decision-making tools should only be developed with extreme caution. These include the risk of misinterpreting decision predicting as decision-making and an unavoidable inherent status quo bias.

\section{Arguments against automatic decision-making}

The push for transparency of the courts is natural and intuitively clear. The courts should be able to judge systematically and predictably and should be accountable in cases when that does not happen. There is a multitude of laws, including the constitution, to make sure that that is the case in countries across the world.

With predictive legal models that become better and better, there appears to be an increasing tendency to evaluate whether judges and their biases may be replaced by machines where those biases can be controlled \cite{chohlas2020blind,khademi2019fairness}. We first would like to note that the idea of having a completely unbiased judge is somewhat peculiar. When we think about laws we think of very elaborate instructions that (try to) account for any situation. And while laws are written in the hope that they would cover most issues, they can often be interpreted in several ways when a new situation is encountered. Legal systems rely on judges interpreting general laws in individual cases \cite{zavrvsnik2020criminal}. By doing so they introduce their personal bias. Different judges might judge the same case differently, and neither of them would be absolutely wrong in their decision. That, of course, does not mean that all judicial biases are justified. For example, discrimination based on gender, skin colour, ethnicity, etc. is of course unacceptable. However, in the high stakes situations we focus on in this paper, some level of bias in judges is inevitable. 

At present, judicial-assisting tools are being introduced and (unfortunately) many of them are being (incorrectly) used as judicial decision-making systems. In the following, we discuss why attempts to build such systems may be misguided in high-stakes situations. 

\subsection{Decision-predicting vs. decision-making systems}

The first issue in discussions on automatic judicial decision-making is that of mixing up definitions. In many tasks that involve machine learning, decision-predicting and decision-making may appear synonymous. However, this is not the case in a court setting.

Classification using machine learning functions by providing the model some kind of representation of (the text of) old cases (i.e.~data points for each case). The model then tries to identify which data points (for example, whether or not the word ''children" occurs in the description of facts available to the court) are most representative for each class (i.e. verdict). Therefore, if there are any biases that can be found within those data points, the system will exploit them to improve its prediction.
However, it may not always be the case that these biases should remain present in the decision-making process \cite{edwards2017slave}. In essence, the decision-predicting system is the one that is able to determine and \textit{distinguish between past decisions}, whereas a decision-making system should be able to \textit{generate new decisions}.

\subsection{Status quo bias}

Given that one always has to train a machine learning system on older cases in order for it to be able to predict the decision of future cases, the system will always reflect the way old cases were decided \cite{campbell2020artificial,berk2019machine}. 
It is no surprise, therefore, that predicting the decisions of future judgements is consistently harder than predicting judgements from the same period \cite{medvedeva2020using}.


Without explicit knowledge about gradually changing concepts (e.g., the introduction of electronic mail), even the most advanced Natural-Language-Processing-based machine learning techniques cannot predict changes in how the law needs to be interpreted. Human intervention is necessary to prevent an inherent status quo bias.



\subsection{Dangers of reverse-engineering}

Another issue when using this type of algorithms lies within the area of cybersecurity. Regarding cybersecurity,  \cite{nichols2019bribing} raised a concern regarding the possibility of hacking and manipulating algorithms in order to benefit self--interested third parties. The author thus argued for transparency in development of algorithms, as have many others \cite{zavrvsnik2020criminal,thomseniudicium,deeks2019machine}. This intuitive argument for transparency, however, also may be problematic. Specifically, a transparent predictive system may be creates an opportunity to abuse the algorithm by using adversarial machine learning techniques \cite{kurakin2016adversarial}.

Consider the following artificial example regarding a low-risk decision-making machine. In the hallway of the court, there is an automatic cookie-vending machine that decides whether or not you can have a cookie. It bases its decision on your personal history. Consequently, the vending machine filled with cookies asks a range of questions to determine whether you can have another cookie, or that you have had enough cookies. To do so, it asks you what kind of cookies you already ate and whether you ate any fruit for breakfast, etc., and it uses stored information from the previous times you have used the cookie machine (similar to risk-assessment tools). You don't know how the machine works, you just answer the questions and unfortunately you are denied a cookie. You think this is unreasonable, as you really wanted a cookie. If you ask for an explanation, and it appears the system is (in legal terms) not explainable, you were denied justice. However, if the model is explainable, it should provide you with the details on how the answers were weighted and how a certain decision was reached. If you know what the system already knows about you and how the system determines its decision, you may be able to create a computer program which provides you with the answers you have to give to increase the number of cookies you are able to obtain. Importantly, if the system is consistent, it is even possible to recreate the algorithm even without access to the specifics of how it works.

Of course, when the risks are higher, the consequences of being able to reverse-engineer a system may be a lot more dire.
Similarly to deceiving the cookie machine, one may be able to play out every scenario before appearing before the court and, for instance, subsequently decide to go to trial or settle.



\section{Discussion and conclusion}

All of the aforementioned issues also exist in judicial systems where machine learning tools are not used. Judges and lawyers deal with precedents to make their cases and come to a certain decision, law firms and advocates try to `reverse-engineer' the judge to predict their behaviour given certain circumstances of the case \cite{bruijn2020upperdogs}. The presence of an automated system, however, amplifies the problem.

In this paper we pointed out the difference between legal decision-predicting and decision-making and argued that the latter has no place in courts where people's quality of life may be at stake. We pointed out that given how machine learning works it is impossible to avoid a status quo bias in decision-making and we stressed that explainable AI is vulnerable to reverse-engineering. If one knows that the machine will judge systematically, given enough data one may be able to predict the outcome of a case in all cases. This may allow law firms to play out various strategies ``in-house" before going to court and therefore evaluating what the results will be. This creates ample opportunities for abusing the system.

In this paper, unfortunately, we do not offer a solution to this issue. We merely caution the reader that although using machine learning has a substantial potential in legal analytics and decision support, we think it should not be introduced for making judicial decisions in situations where human rights are at stake. In addition, in cases where it has already been introduced, stricter regulations need to be enforced to make sure that decisions are never made solely on the basis of a machine learning system's predictions.
%
%
%
 \bibliographystyle{splncs04}
 \bibliography{bib}

\begin{thebibliography}{10}
\providecommand{\url}[1]{\texttt{#1}}
\providecommand{\urlprefix}{URL }
\providecommand{\doi}[1]{https://doi.org/#1}

\bibitem{aletras2016predicting}
Aletras, N., Tsarapatsanis, D., Preo{\c{t}}iuc-Pietro, D., Lampos, V.:
  Predicting judicial decisions of the {E}uropean {C}ourt of {H}uman {R}ights:
  A natural language processing perspective. PeerJ Computer Science  \textbf{2}
  (2016)

\bibitem{angwin2016machine}
Angwin, J., Larson, J., Mattu, S., Kirchner, L.: Machine bias: there’s
  software used across the country to predict future criminals. and it’s
  biased against blacks. {P}ro{P}ublica 2016 (2016),
  https://www.propublica.org/article/machine-bias-risk-assessments-in-criminal-sentencing

\bibitem{berk2019machine}
Berk, R., Berk, Drougas: Machine learning risk assessments in criminal justice
  settings. Springer (2019)

\bibitem{bex2020juridische}
Bex, F., Prakken, H.: De juridische voorspelindustrie: onzinnige hype of
  nuttige ontwikkeling? Ars aequi  \textbf{69},  255--259 (2020)

\bibitem{bruijn2020upperdogs}
Bruijn, L.M., Vols, M.: Upperdogs versus underdogs. Recht der Werkelijkheid
  \textbf{1},  25--49 (2020)

\bibitem{campbell2020artificial}
Campbell, R.W.: Artificial intelligence in the courtroom: The delivery of
  justice in the age of machine learning. Colo. Tech. LJ  \textbf{18}, ~323
  (2020)

\bibitem{chalkidis-etal-2019-neural}
Chalkidis, I., Androutsopoulos, I., Aletras, N.: Neural legal judgment
  prediction in {E}nglish. In: Proceedings of the 57th Annual Meeting of the
  Association for Computational Linguistics. pp. 4317--4323. Association for
  Computational Linguistics, Florence, Italy (Jul 2019).
  \doi{10.18653/v1/P19-1424}, \url{https://www.aclweb.org/anthology/P19-1424}

\bibitem{chen-etal-2019-charge}
Chen, H., Cai, D., Dai, W., Dai, Z., Ding, Y.: Charge-based prison term
  prediction with deep gating network. In: Proceedings of the 2019 Conference
  on Empirical Methods in Natural Language Processing and the 9th International
  Joint Conference on Natural Language Processing (EMNLP-IJCNLP). pp.
  6363--6368. Association for Computational Linguistics, Hong Kong, China (Nov
  2019). \doi{10.18653/v1/D19-1667},
  \url{https://www.aclweb.org/anthology/D19-1667}

\bibitem{chohlas2020blind}
Chohlas-Wood, A., Nudell, J., Nyarko, J., Goel, S.: Blind justice:
  Algorithmically masking race in charging decisions. Tech. rep., Technical
  report (2020)

\bibitem{deeks2019machine}
Deeks, A., Lubell, N., Murray, D.: Machine learning, artificial intelligence,
  and the use of force by states. J. Nat'l Sec. L. \& Pol'y  \textbf{10}, ~1
  (2019)

\bibitem{dressel2018accuracy}
Dressel, J., Farid, H.: The accuracy, fairness, and limits of predicting
  recidivism. Science advances  \textbf{4}(1),  55--80 (2018)

\bibitem{edwards2017slave}
Edwards, L., Veale, M.: Slave to the algorithm: Why a right to an explanation
  is probably not the remedy you are looking for. Duke L. \& Tech. Rev.
  \textbf{16}, ~18 (2017)

\bibitem{khademi2019fairness}
Khademi, A., Lee, S., Foley, D., Honavar, V.: Fairness in algorithmic decision
  making: An excursion through the lens of causality. In: The World Wide Web
  Conference. pp. 2907--2914 (2019)

\bibitem{kurakin2016adversarial}
Kurakin, A., Goodfellow, I., Bengio, S.: Adversarial machine learning at scale.
  arXiv preprint arXiv:1611.01236  (2016)

\bibitem{medvedeva2020using}
Medvedeva, M., Vols, M., Wieling, M.: Using machine learning to predict
  decisions of the {E}uropean {C}ourt of {H}uman {R}ights. Artificial
  Intelligence and Law  \textbf{28}(2),  237--266 (2020)

\bibitem{nichols2019bribing}
Nichols, P.M.: Bribing the machine: Protecting the integrity of algorithms as
  the revolution begins. American Business Law Journal  \textbf{56}(4),
  771--814 (2019)

\bibitem{niiler2019can}
Niiler, E.: Can {AI} be a fair judge in court. estonia thinks so. WIRED
  (2019), www.wired.com/story/can-ai-be-fair-judge-court-estonia-thinks-so/

\bibitem{stevenson2018assessing}
Stevenson, M.: Assessing risk assessment in action. Minn. L. Rev.
  \textbf{103}, ~303 (2018)

\bibitem{stranieri2006knowledge}
Stranieri, A., Zeleznikow, J.: Knowledge discovery from legal databases,
  vol.~69. Springer Science \& Business Media (2006)

\bibitem{thomseniudicium}
Thomsen, F.: Iudicium ex machinae--the ethical challenges of automated
  decision-making in criminal sentencing

\bibitem{zavrvsnik2020criminal}
Zavr{\v{s}}nik, A.: Criminal justice, artificial intelligence systems, and
  human rights. In: ERA Forum. vol.~20, pp. 567--583. Springer (2020)

\bibitem{zhong2018legal}
Zhong, H., Guo, Z., Tu, C., Xiao, C., Liu, Z., Sun, M.: Legal judgment
  prediction via topological learning. In: Proceedings of the 2018 Conference
  on Empirical Methods in Natural Language Processing. pp. 3540--3549 (2018)

\end{thebibliography}

\end{document}